# FAST SINGLE-SHOT SHIP INSTANCE SEGMENTATION BASED ON POLAR TEMPLATE MASK IN REMOTE SENSING IMAGES

*Zhenhang Huang, Shihao Sun, Ruirui Li\**

College of Information Science & Technology, Beijing University of Chemical Technology

ilydouble@gmail.com

## ABSTRACT

Object detection and instance segmentation in remote sensing images is a fundamental and challenging task, due to the complexity of scenes and targets. The latest methods tried to take into account both the efficiency and the accuracy of instance segmentation. In order to improve both of them, in this paper, we propose a single-shot convolutional neural network structure, which is conceptually simple and straightforward, and meanwhile makes up for the problem of low accuracy of single-shot networks. Our method, termed with *SSS-Net*, detects targets based on the location of the object's center and the distances between the center and the points on the silhouette sampling with non-uniform angle intervals, thereby achieving a balanced sampling of lines in mask generation. In addition, we propose a non-uniform polar template IoU based on the contour template in polar coordinates. Experiments on both the Airbus Ship Detection Challenge dataset and the ISAID-ships dataset show that *SSS-Net* has strong competitiveness in precision and speed for ship instance segmentation.

## 1. INTRODUCTION

Ship detection and recognition in high-resolution aerial images are challenging tasks, which play a vital role in many related applications, e.g., maritime security and port resources management. In recent years, with the fast development of deep learning techniques, the accuracy of ship detection has been improved a lot. As a result, higher demands are raised to extract the segmentation mask of ship instances.

Fig.1 shows the result forms of ship detection, ship segmentation and ship instance segmentation. Differently, instance segmentation is not only dedicated to predicting each pixel to different class label, but also distinguishing different objects of the same category. The task of instance segmentation is beneficial to capture the location, contour, and correlation information of the objects. It goes without saying that in the field of remote sensing[6], ship instance segmentation is the fundamental step for further ship recognition and relevant applications.

Extending Faster-RCNN[5], Mask-RCNN[2] is a powerful network structure for instance segmentation by adding a branch for predicting an object mask in parallel.

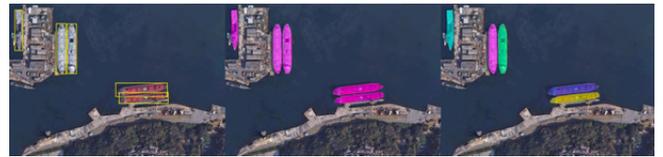

(a) detection   (b) segmentation   (c) instance

**Fig.1.** (a), (b) and (c) visualize the result forms of ship detection, ship segmentation and ship instance segmentation, respectively.

But in our task of ship instance segmentation, there are still some problems. In maritime aerial images, for a ship with a large aspect ratio, the redundancy region within the proposal will be relatively large, which is unfavorable to the operation of non-maximum suppression (NMS), often resulting in missing detection. RRPN[8] uses rotation anchors to relieve the problem, but causing information loss when processing the ROI. In real life, ships in the harbor are always densely docked. To avoid confusion between the objects. SLCMASK-Net[6] introduced a sequence local context module. However, all above attempts are two-stage approaches largely depending on the region proposals, which achieve high accuracy but suffer from slow run-times that make them infeasible for many real-world applications. In contrast, proposal-free instance segmentation method has received increasing attention due to its concise and efficient pipeline, enabling real-world applications with substantial efficiency requirements such as maritime drone rescue. Generally, proposal-free methods generate instance-agnostic semantic segmentation labels and instance-aware features to group pixels into different object instances. SSAP[3] introduces a pixel-pair affinity pyramid to improve instances generation. It not only achieves state-of-the-art results on Cityscape street view images, but also gets 5× speedup. PolarMask[1] is a simple, fully convolutional instance segmentation framework. It formulates instance segmentation problem as instance center classification and dense distance regression in a polar coordinate. It is flexible and able to achieve competitive accuracy.

Inspired by its conciseness and efficiency, the proposed SSS-Net is primarily based on PolarMask[1]. We are the first to try segment ship instances using PolarMask architecture. In the experimental observation, PolarMask sometimes fails when dealing with ship contours. It is because that ships in the maritime remote sensing images are relatively too small and are easy to lose detection. Moreover, ships have shuttle-shaped silhouette with large aspect ratio, resulting in ineffective and inefficient using uniform angular sampling in the polar regression head.

In view of the above problems, we propose a non-uniform angular sampling method to model the ship silhouette in polar coordinate system. It is called polar template mask (PTM). In summary, we made the following contributions to the community:

- An anchor-free, single-shot instance segmentation method for ship instance segmentation in remotely sensed maritime images, which is fast and accurate.
- Ship template mask as well as non-uniform polar IoU loss in the mask regression branch to better model the ship silhouette in the polar coordinate.

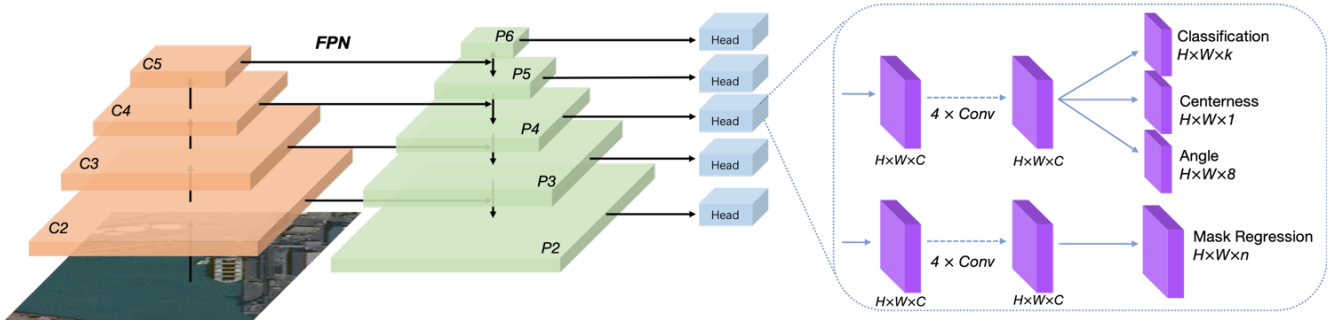

**Fig.2.** The architecture of our proposed SSS-Net. The left part is the backbone for feature extraction, which is connected to the Feature Pyramid Network to generate multi-scale feature maps. The right part composed of two branches. The upper side is the Classification, Centerness and Angle branch for predicting instance categories, centerness and main angles, while the lower side is the Mask Regression branch for generating instance mask distances.

## 2. PROPOSED METHOD

As shown in Fig.2, *SSS-Net* is based on PolarMask[1], which is a concise structure with fast prediction speed. *SSS-Net* is an end-to-end, unified network that includes a backbone network module, a feature pyramid network(FPN) [4] module, and prediction heads for dual subtasks. One task is used to predict the category, the center position, and the main direction angle of the instances. The other is used to predict the mask. Just through one-time inference, it can get segmented images containing predicted instances.

### 2.1. Polar Template Mask

Ships in remotely sensed images commonly have a large aspect ratio, as shown in Fig.3(a). The masks generated by sampling lines with equivalent interval angles (Fig.3(c)) are not accurate and smooth enough in the two ends of the ship. As a result, it wastes a lot of computing resources in the horizontal directions.

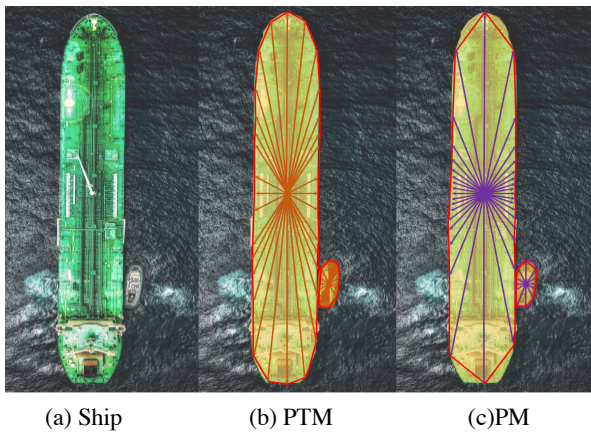

(a) Ship     (b) PTM     (c)PM

**Fig.3.** Comparison of polar representations between the proposed polar template mask(b) and PolarMask(c).

Facing this problem, An intuitive solution is to increase the sampling density of the two ends while reducing the others, which is shown in Fig.3(b). We call this mask representation method as Polar Template Mask.

To be specific, given an instance mask, we first get the mass center $(x_c, y_c)$ of the instance and the points on the contour $(x_i, y_i)$, $i = 1, 2, \cdots, N$. Starting from the center, $n$ rays are emitted based on the template. We obtain the new contour $(\rho_i, \theta_i)$, $i = 1, 2, \cdots, n$. Where $\rho$ donates the distance from mass center to the new contour, and $\theta$ donates the vector tilt angle.

In our method, we define the direction of the longest ray as the main direction. After that the instance mask is divided into 8 fan-shaped areas on the basis of the main direction, and then subdivided. The two areas pointed by the main direction and their symmetrical areas are subdivided into 4*M* pieces, the remaining areas are subdivided into *M* pieces. (We set *M*=2 in this experiment) Finally, a 20*M* polygon is formed. Fig.4 shows the simplified Polar Template Mask.

Since the angle intervals are predefined by template, only the lengths of the rays need to be predicted. In this way, we formulate the instance segmentation as instance classification, template classification and distances regression.

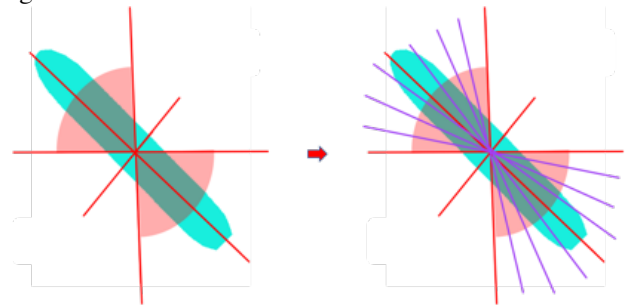

**Fig.4.** Polar Template Mask.

## 2.2. Polar Centerness

Centerness is important for ship segmentation. It is used to suppress the low quality detected ships. We directly transfer the polar centerness into our scheme. Given a set $\{d_1, d_2, ..., d_n\}$ for the lengths of the n rays of one instance, where $d_{\max}$ and $d_{\min}$ are the maximum and minimum of the set. The polar centerness is as follows:

$$Polar\ Centerness = \sqrt{\frac{min(d_1, d_2, ..., d_n)}{max(d_1, d_2, ..., d_n)}}$$

## 2.3. Loss Function

Assuming $p_{x,y}$ is the classification score, $O_{x,y}$ is the predicted centerness, $d_{x,y}$ is the regression prediction for each location on the feature maps and $d^*_{x,y}$ is the correspondent value of the ground truth. In addition, $A_{x,y}$ represents the predicted main angle and $A^*_{x,y}$ represents the corresponding real angle. The overall loss function is defined as follows:

$$L = \frac{1}{N_{pos}} \sum_{x,y} L_{cls}(p_{x,y}, c^*_{x,y}) + \frac{1}{N_{pos}} \sum_{x,y} \mathbb{1}_{\{c^*_{x,y}>0\}} L_{polar}(d_{x,y}, d^*_{x,y})$$

$$+ \frac{1}{N_{pos}} \sum_{x,y} L_{center}(O_{x,y}, O^*_{x,y}) + \frac{1}{N_{pos}} \sum_{x,y} L_{angle}(A_{x,y}, A^*_{x,y})$$

Where $L_{cls}$ is the Focal Loss[10], $L_{polar}$ is the Polar Template IoU Loss which will be described in detail later, $L_{center}$ and $L_{angle}$ are the binary cross entropy loss of centerness and main angles. $N_{pos}$ denotes the number of positive samples. The summation is calculated over all locations on the feature maps. $\mathbb{1}_{\{c^*_{x,y}>0\}}$ is the indicator function, being 1 if $c^*_{x,y} > 0$ and 0 otherwise.

To be specific, we proposed Polar Template IoU Loss is the binary cross entropy (BCE) loss of Polar Template IoU. According to the definition of IoU, the polar template IoU is converted to:

$$PT\ IoU\ Loss = \log \frac{\sum_{i=1}^{N} max(d, d^*) \Delta\theta_i}{\sum_{i=1}^{N} min(d, d^*) \Delta\theta_i}$$

Where $\Delta\theta_i$ represents the angle between vectors $d_i$ and $d_{i+1}$. The Polar Template IoU Loss can be well applied in the *SSS-Net* scenario because of its differentiability, which can achieve back propagation and the overall prediction regression target, which can greatly improve the overall performance.

## 3. EXPERIMENTS AND RESULTS

In this section, we evaluate our method on the *Airbus Ship* and the *ISAID* dataset and compare the results to state-of-the-art methods, adopting the official F2 score and the AP (Average Precision) evaluation metrics. Additionally, we also visualize the results for further analysis.

### 3.1. Dataset Description

To validate the effectiveness of the proposed method, we conduct extensive experiments on two publicly accessible datasets: Airbus Ship Detection[7] and ISAID[9].

**Airbus Ship Detection.** The dataset consists of 208,172 images (192,566 for training, 15,606 for testing). There are 42,566 valid images only containing ship annotation(including 81,689 ship instances). The image scale of the dataset is about 768 × 768 pixels. In addition, we divided the official training set into a new training set(85%) and a testing set(15%), which helps to evaluate our method offline.

**ISAID.** The dataset consists of 2,348 images (1,411 for training, 917 for validation, 937 for testing) with annotations on 15 categories of objects with multiple resolutions (800 × 800 through 4,000 × 4,000 pixels). We divided these large images into 30,429 pictures with a uniform 768 × 768 resolution, containing 108,383 ship instances for training and 31,137 ship instances for validation. The validation dataset is used as the testing set to evaluate our method.

### 3.2. Experimental Results and Analysis

Tab.1 shows the results officially submitted in the Airbus Ship Detection Challenge. It can be seen that the Mask-RCNN outperforms in instance segmentation task. The Polarmask is lower than it by about 25% in both public and private test scores. However, it is surprising that our method can achieve the accuracy close to the two-stage method under the F2 score. It illustrates the superiority of the method in ship segmentation.

Tab.2 describes in detail the performance of various methods on the airbus ship dataset on the currently recognized the evaluation metric AP. It can be seen that by deepening the backbone, the detection accuracy greatly is improved, meanwhile with a sharp increase of FLOPs and parameter amounts. Moreover, Mask-RCNN still outperform in the AP with large parameter amounts. But our SSS-Net, only adding negligible parameter amounts, can improve the AP by about 2.6%. In addition, it should be noted that the factor leading to the low accuracy of the single-shot method is the low APS (the small targets AP). And the number of small targets accounts for a large proportion in remotely sensed images.

We also conducted experiments in ISAID dataset, another authoritative remote sensing dataset. The results are shown in Tab.3. Note that the overall performance dropped significantly compared with the Airbus Ship. This may be due to the large proportion of small and dense targets. Therefore, we count the small, medium, and large targets of the above two datasets(according to the COCO rules). The ratios are about 5.6:3.1:1 and 30.7:13.4:1, indicating the extreme imbalance in the ship size distribution in remote sensing images. Additionally, we evaluate the inference speed of different methods. As shown in Tab.4, SSS-Net performed slightly better than PolarMask and nearly 30% faster than Mask-RCNN.

Fig.5 shows the comparison of the visualization results. We selected several representative pictures. The first row is selected to represent the ships close to the land, the port and moored along the coast. The second row represents the ship with a boat or with a tail independent in the sea. It can be seen that in the first row, the masks of PolarMask are rough. Mask-RCNN incorrectly recognizes the upper left corner of the land as a ship. In the second row, single-shot methods are more difficult to detect for small ships with tailing. In the third row, we found the insufficient segmentation of PolarMask and the over-segmentation of Mask-RCNN, which may be attributed to the rough labeling.

**Tab.2.** Segmentation performance on Airbus Ship of Mask-RCNN, PolarMask and SSS-Net.

| Backbone | Method | $AP$ | $AP_{50}$ | $AP_{75}$ | $AP_S$ | $AP_M$ | $AP_L$ | $FLOPs$ | $Params$ |
|---|---|---|---|---|---|---|---|---|---|
| ResNet-50 | Mask-RCNN | 60.4 | 86.9 | 71.4 | 37.5 | 76.1 | 86.4 | 273.58 | 43.75 |
| | PolarMask | 44.8 | 77.2 | 50.6 | 20.1 | 63.0 | 67.7 | 248.69 | 34.28 |
| | Our Method | 47.4 | 78.9 | 54.1 | 23.6 | 66.0 | 68.6 | 248.89 | 34.29 |
| ResNet-101 | Mask-RCNN | 67.8 | 88.3 | 79.4 | 48.9 | 81.7 | 87.4 | 463.53 | 64.06 |
| | PolarMask | 52.3 | 81.7 | 60.5 | 28.7 | 70.5 | 72.9 | 324.76 | 53.22 |
| | Our Method | 54.9 | 87.8 | 63.5 | 32.7 | 71.3 | 74.4 | 324.96 | 53.23 |

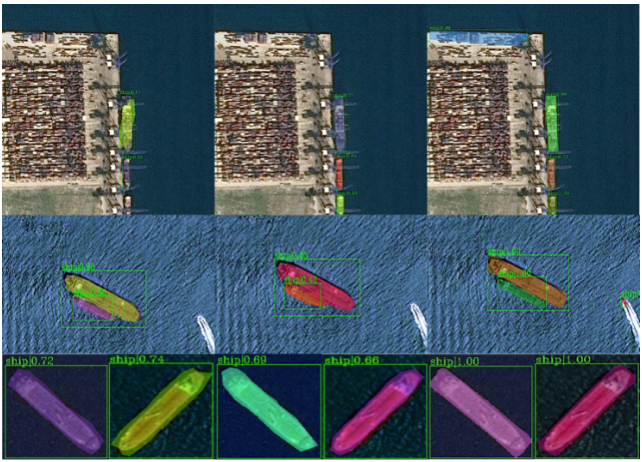

(a) PolarMask  (b) SSS-Net  (c) Mask-RCNN

**Fig.5.** The visualizations of ship instance segmentation.

**Tab.1.** Instance segmentation F2 score.

| F2 Score | Mask-RCNN | PolarMask | SSS-Net |
|---|---|---|---|
| Private Score | 81.4 | 55.5 | 80.0 |
| Public Score | 66.3 | 43.7 | 64.2 |

**Tab.3.** Segmentation AP performance of ISAID dataset.

| Backbone | Method | $AP$ | $AP_{50}$ | $AP_{75}$ |
|---|---|---|---|---|
| ResNet-50 | Mask-RCNN | 43.1 | 64.9 | 53.6 |
| | PolarMask | 34.9 | 62.0 | 38.0 |
| | Our Method | 35.5 | 62.2 | 39.5 |
| ResNet-101 | Mask-RCNN | 43.8 | 65.6 | 53.7 |
| | PolarMask | 35.2 | 61.7 | 39.2 |
| | Our Method | 36.2 | 63.0 | 40.9 |

**Tab.4.** The speed of instance segmentation inference.

| FPS | Mask-RCNN | PolarMask | SSS-Net |
|---|---|---|---|
| ResNet-50 | 10.8 | 15.6 | 16.0 |

(The results are reported on an RTX2080Ti.)

## 4. CONCLUSION

In this paper, we are committed to improving the imbalance problem of segmentation efficiency and accuracy in remotely sensed ship images. We propose a more accurate, faster, and single-shot network structure, SSS-Net, which is based on a novel mask representation. And we adopt a non-uniform angle template sampling algorithm to effectively reduce redundant information, and refine the silhouette. In addition, this network structure can still be easily embedded in a two-stage segmentation network, which deserves further study.

## 5. REFERNCES